# AR-GAN: Generative Adversarial Network-Based Defense Method Against Adversarial Attacks on the Traffic Sign Classification System of Autonomous Vehicles

Generative Adversarial Network-based Resilient Traffic Sign Classification System

M Sabbir Salek, M.S., Salek*

Ph.D., Glenn Department of Civil Engineering, Clemson University, Clemson, SC 29634, USA, msalek@clemson.edu

Abdullah Al Mamun, A.A., Mamun

Ph.D., Glenn Department of Civil Engineering, Clemson University, Clemson, SC 29634, USA, abdullm@clemson.edu

Mashrur Chowdhury, M., Chowdhury

Ph.D., Professor, Glenn Department of Civil Engineering, Clemson University, Clemson, SC 29634, USA, mac@clemson.edu

Autonomous vehicles (AVs) rely on deep neural network (DNN)-based classification systems to recognize traffic signs. However, DNN models are vulnerable to adversarial attacks that can cause misclassification by introducing slight perturbations to an input image, the consequence of which can be fatal for AVs. This study developed a generative adversarial network (GAN)-based defense method for traffic sign classification in an AV, referred to as the attack-resilient GAN (AR-GAN). The novelty of the AR-GAN lies in (i) assuming zero knowledge of adversarial attack models and samples and (ii) providing consistently high traffic sign classification performance under various adversarial attack types. The AR-GAN classification system consists of a generator that denoises an image by reconstruction, and a classifier that classifies the reconstructed image. The authors have tested the AR-GAN under no-attack and under various adversarial attacks, such as Fast Gradient Sign Method (FGSM), DeepFool, Carlini and Wagner (C&W), and Projected Gradient Descent (PGD). The authors considered two forms of these attacks, i.e., (i) black-box attacks (assuming the attackers possess no prior knowledge of the classifier), and (ii) white-box attacks (assuming the attackers possess full knowledge of the classifier). The classification performance of the AR-GAN was compared with several benchmark adversarial defense methods. The results showed that both the AR-GAN and the benchmark defense methods are resilient against black-box attacks and could achieve similar classification performance to that of the unperturbed images. However, for all the white-box attacks considered in this study, the AR-GAN method outperformed the benchmark defense methods. In addition, the AR-GAN was able to maintain its high classification performance under varied white-box adversarial perturbation magnitudes, whereas the performance of the other defense methods dropped abruptly at increased perturbation magnitudes.

## 1 INTRODUCTION

### 1.1 Background and Motivation

Autonomous vehicles (AVs) perform the autonomous driving task with the help of a suite of sensors and software. Sensors, including camera, light detection and ranging (LiDAR), and radio detection and ranging (Radar), help AVs perceive their surrounding environment [24]. The sensed data is fed to the AV perception module, where the relevant information for autonomous navigation is extracted, such as traffic signs, signals, lane markings, surrounding vehicles, pedestrians, and obstacles. Nowadays, even many human-driven vehicles have dashboard camera-based traffic sign classification systems. These systems are typically highly dependent on machine learning (ML) or deep learning (DL) models, especially deep neural networks (DNNs) [37]. However, since AVs rely on such systems to realize roadway regulations and maneuver accordingly, compromised information regarding roadway traffic signs can be hazardous for AVs. Thus, researchers have emphasized developing DNN-based accurate traffic sign classification systems over the past few years [19].

---

* Corresponding author.

However, DNN-based classification systems have some cybersecurity vulnerabilities. For example, an adversarial attack can introduce slight perturbations to the input images fed to a traffic sign classification system and cause the underlying DNN models to misclassify the signs on the roadway. These perturbations can be so minimal that they are imperceptible to regular human eyes. However, they can be effective in deceiving the DNN models used in AVs' traffic sign classification systems. To this end, the authors aim to develop an AV traffic sign classification system resilient to such adversarial attacks.

Adversarial attacks can be categorized based on the extent of knowledge the attack model or the attacker has about the DNN models responsible for classification [13]. If the attacker has no knowledge about the DNN-based classification models, e.g., its architecture and parameters, then it is called a black-box attack. If the attacker has partial knowledge about the DNN-based classification models, e.g., the attacker may know the DNN architecture but may be unaware of its trained parameters, then it is known as a gray-box attack. If the attacker has full knowledge of the DNN-based classification models, then the attack is called a white-box attack. In a white-box attack, the victim, i.e., in this case, the DNN-based classification system, is at a maximum disadvantage. This is because the attacker can craft an adversarial attack in such a way that the DNN-based classification system and its defense methods remain utterly unaware of the attack. In this study, the goal is to develop an AV traffic sign classification system that is resilient to adversarial attacks, particularly to white-box attacks. In a connected and AV environment, it is reasonable to assume that an attacker could access sensitive system configurations or algorithms even without physically accessing the AV.

Different defense methods have been proposed by researchers over the past few years to protect image classification systems from adversarial attacks [13], such as modification of the DNNs [29,33], adversarial training [2], input transformation [21,38], and input reconstruction [11,17]. The recent breakthroughs in generative adversarial networks (GANs) have opened opportunities to utilize GANs for defense against adversarial attacks. For example, Samangouei et al. [36] introduced a Wasserstein GAN (WGAN)-based defense method, known as the Defense-GAN, which can protect image classification systems against known and unknown adversarial attacks by reconstructing the input images before feeding them to a classifier. The generator model in the Defense-GAN method was trained to generate samples similar to the unperturbed (legitimate) images given a random input latent vector. The random input latent vector is found by solving an optimization problem to minimize the reconstruction error of the generator. However, WGAN is known for suffering from issues associated with weight clipping, such as vanishing gradient and non-convergence of the discriminator, which makes the training of an appropriate WGAN model very difficult [8]. In [11], Jin et al. developed a defense method for image classification systems against adversarial attacks called the Adversarial Perturbation Elimination with GAN (APE-GAN). The generator of the APE-GAN was trained with adversarial examples to eliminate adversarial perturbations by making changes to the input images. However, the authors in [11] also utilized the loss function of WGAN, which inherits the same issues as discussed earlier. Besides, adversarial training-based defense methods work well for known attacks only and are susceptible to unknown attacks [40]. Laykaviriyakul and Phaisangittisagul [17] presented an adversarial defense framework for image classification systems based on the DiscoGAN architecture [16], where the authors utilized an attacker model to create adversarial examples from the training data and a defender model to reconstruct unperturbed images from the adversarial images. These two models were trained in tandem to play a competitive game with each other. However, the authors in [17] also used an adversarial training-based defense approach, which does not ensure classification performance under unknown attacks on which the models were not trained. Besides, all these studies considered benchmark datasets like CIFAR-10, MNIST, and Fashion MNIST. Thus, their performance on real-world datasets is not explored yet.

In this study, the authors' aim is to develop an attack-resilient GAN-based defense method for an AV traffic sign classification system, which the authors refer to as the AR-GAN, that can protect the perception module of an AV from unknown attacks. The authors used a WGAN-based loss function with gradient penalty (WGAN-GP) to train the GAN models, which was shown to overcome common issues with GAN/WGAN training. The authors trained the GAN models and classifiers on the unperturbed traffic sign images only so that all types of adversarial attacks are unknown to the models. For evaluation, the authors considered both black-box and white-box attacks with varied perturbation magnitudes.

## 1.2 Contribution

Much work has been done on DNN-based traffic sign classification systems for AVs in the literature [37,19,34]. However, to the best of the authors' knowledge, none of the existing studies utilized a GAN-based adversarial defense method for AV traffic sign classification systems. In this study, the authors developed an adversarial attack-resilient traffic sign classification system based on GAN (AR-GAN) for AVs, which is designed to be robust against both black-box and white-box adversarial attacks without requiring any prior knowledge about the attack models. The AR-GAN method utilizes a WGAN-GP-based loss function to overcome typical convergence issues with GANs, such as mode collapse and vanishing gradient. The generator in the AR-GAN method is based on the deep convolutional GAN (DCGAN) architecture and trained to generate unperturbed samples from adversarial samples before feeding them to the classifier. The classifier in the AR-GAN method is based on the residual network (ResNet) architecture [10] and trained on traffic sign images reconstructed by the generator. In addition, the AR-GAN uses a particular training framework to ensure the performance of the models used in the AR-GAN traffic sign classification system. Thus, the generator and the classifier in the AR-GAN traffic sign classification system can achieve similar and high traffic sign classification accuracies under no-attack condition, as well as under various black-box and white-box adversarial attack scenarios. The authors evaluated the AR-GAN traffic sign classification system's classification performance and resiliency against adversarial attacks with a real-world traffic sign dataset in this study. Also, the AR-GAN traffic sign classification system can provide consistently high classification performance under different perturbation magnitudes, unlike the traditional adversarial defense methods.

## 2 LITERATURE REVIEW

Significant progress has been made in traffic sign classification in recent years, with a growing focus on using DNN-based techniques. In this section, the authors present some notable contributions in this area including the studies that focused on developing defense methods for AV traffic sign classification systems.

Kerim and Efe [12] developed a hybrid neural network (NN) to classify traffic signs using various features, including Histograms of Oriented Gradients (HOG) and a combination of color, HOG, and Local Binary Patterns (LBP), with an accuracy of 95%. They used the German Traffic Sign Recognition Benchmark (GTSRB) and the Chinese Traffic Sign Recognition Dataset (TSRD), and applied data augmentation to enhance model performance. Kheder and Mohammed [15] improved the traditional LeNet-5 convolutional neural network (CNN) model architecture by adding layers and integrating image preprocessing algorithms to enhance performance. Their model achieved 99.12% accuracy on the GTSRB and 99.78% on the extended GTSRB (EGTSRB) datasets. Panduarangan et al. [28] used preprocessing techniques, such as median filtering and histogram equalization, and applied ML and DL algorithms, including Extreme Learning Machine (ELM), Linear Discriminant Analysis (LDA), Principal Component Analysis (PCA), and CNN-General Regression NN (GRNN), to develop a traffic sign recognition model. They achieved an accuracy of 99.41% on the GTSRB dataset.

In addition to the studies mentioned above, many others developed DNN architectures to classify traffic signs using various datasets [19,37]. However, only a few studies have developed DNN-based defense methods to improve the resilience and robustness of traffic sign classification systems against adversarial attacks. Among them, Li et al. [18] proposed a defense method using a spatial transformation module to counter adversarial attacks, showing an average accuracy of 73.95% on GTSRB dataset when subjected to untargeted white-box Fast Gradient Sign Method (FGSM) attacks. Hashemi et al. [9] developed a cost function, Regularized Guided Complement Entropy (RGCE), improving robustness against various adversarial attacks and maintaining performance on clean images. The RGCE recorded accuracies of 90.24% and 74.44% when targeted by the FGSM and the Projected Gradient Descent (PGD) L-infinite norm attacks, with a perturbation magnitude of 0.04. The adversarial samples in [9] were generated using a ResNet18 classifier trained on the GTSRB dataset. Khan et al. [14] developed a DNN-based hybrid defense method based on Inception-V3 and ResNet152 DNN models and incorporated random filtering, ensembling, and local feature mapping defense methods. Their proposed defense method's evaluation on a modified subset version of the extended LISA traffic sign database showed 99% classification accuracy on average in the absence of attacks and 88% classification accuracy on average against various adversarial attacks, such as FGSM, Momentum Iterative Method (MIM), PGD, and Carlini and Wagner (C&W) attacks. The hybrid defense method in [14] was reported to perform better than some other traditional defense methods, such as feature squeezing, JPEG filtering, binary filtering, and random filtering. Majumder et al. [23] developed hybrid classical-quantum DL models using pre-trained ResNet18 CNN and quantum gates in the classical and quantum layers, respectively. They evaluated two hybrid models on a modified subset of the LISA dataset. One of the hybrid models outperformed the classical one under PGD attack but underperformed against FGSM attack.

Although the abovementioned studies proposed different DNN-based traffic sign classification systems for defending against adversarial attacks, none considered a generative NN-based method. Recent advancements in GANs present a new opportunity for AV applications, such as AV traffic sign classification, which is the focus of this study.

## 3 ATTACK MODELS

Numerous adversarial attack models have been developed by researchers in the past few years to target NN-based classifiers [13]. In this study, the authors considered both black-box and white-box attacks. In both cases, the attacker aims to find a perturbation $\delta$ that will cause misclassification by the classifier when added to legitimate input $x \in \mathbb{R}^n$, i.e., $\tilde{x} = x + \delta$, where $\tilde{x}$ is the modified input or the adversarial example that may cause misclassification. In this section, the authors discuss the various adversarial attack models used in this study.

### 3.1 Fast Gradient Sign Method (FGSM) Attack

FGSM is a simple but effective adversarial attack proposed by Goodfellow et al. [5]. FGSM utilizes the gradient of the loss (cost) function to generate adversarial examples. Despite its simplicity, FGSM is one of the widely popular adversarial attacks due to its effectiveness in causing misclassification with high confidence [5]. Given an input image $x$ and a classifier with parameters $\theta$, FGSM attack aims to generate an adversarial example $\tilde{x} = x + \delta$, where the added perturbation $\delta$ is determined by computing the gradient of the loss function with respect to the input $x$ as follows [5],

$$\delta = \varepsilon.sign(\nabla_x J(\theta, x, y)) \qquad (1)$$

where, $J$ denotes the loss function, $y$ denotes the output class, $\nabla_x$ is a differential operator with respect to $x$, and $\varepsilon$ denotes the magnitude of perturbation chosen by the attacker.

### 3.2 DeepFool Attack

DeepFool is another simple yet effective optimization-based iterative adversarial attack model proposed by Moosavi-Dezfooli et al. [26], which was reported to be more effective than the FGSM attack on MNIST and CIFAR-10 datasets. Given a binary classifier model, this attack aims to find the minimum perturbation $\delta^*$ that would cause misclassification by shifting the input $x$ to the other side of the decision boundary. The minimum perturbation $\delta^*$ is determined through an optimization problem as follows [26],

$$\delta^* = arg\min_{\delta} \|\delta\|_2 \tag{2}$$

$$\text{subject to: } sign(f(x+\delta)) \neq sign(f(x))$$

where, $f$ is an arbitrary binary classification model. At $i^{th}$ iteration, DeepFool updates $\delta$ by linearizing the classification boundary around the current point $x_i$. To ensure that the final perturbation $\hat{\delta}$ crosses the decision boundary and causes misclassification, $\delta^*$ is multiplied by a constant $(1 + \eta)$, where $\eta \ll 1$. The authors in [26] extended this approach to multi-class classifiers as well.

### 3.3 Carlini and Wagner (C&W) Attack

Carlini and Wagner [3] introduced an optimization-based iterative adversarial attack known as the C&W attack. The C&W attack is a powerful attack that has been reported to cause very low classification accuracy on benchmark datasets, such as MNIST and CIFAR datasets [3]. In a C&W attack, given an input image $x \in \mathbb{R}^n$, the goal of the attack is to determine an optimal perturbation $\delta^*$ to yield misclassification of the adversarial example $\tilde{x} = x + \delta^*$ by a target classifier. In an $\ell_2$-norm C&W attack, the optimal perturbation $\delta^*$ is determined through the following optimization problem [3],

$$\delta^* = arg\min_{\delta} \|\delta\|_2 + c.f(x+\delta) \tag{3}$$

$$\text{subject to: } x + \delta \in [0, 1]^n$$

where, $\|.\|_2$ denotes the $\ell_2$-norm, $c > 0$ is an arbitrary constant, and $f$ is an objective function that helps the misclassification, which is chosen based on the knowledge of the target classifier model. Apart from the $\ell_2$-norm attack explained above, C&W attacks can also be performed using $\ell_0$ and $\ell_\infty$ norms.

### 3.4 Projected Gradient Descent (PGD) Attack

PGD is a powerful adversarial attack model that also utilizes an optimization-based iterative approach. The authors in [22] showed the effectiveness of PGD attacks on MNIST and CIFAR-10 datasets, in which PGD was able to yield lower classification accuracy on the datasets compared to FGSM and C&W attacks. Given an input image $x$, a classifier with parameters $\theta$, and a perturbation $\delta$ to be optimized for a PGD attack, the optimization problem can be written as [22],

$$\delta^* = arg\min_{\delta} \|\delta\|_2 \tag{4}$$

$$\text{subject to: } x + \delta \in [x_{min}, x_{max}]$$

where, $\|.\|_2$ denotes the $\ell_2$-norm, and $x_{min}$ and $x_{max}$ are the minimum and the maximum values of each pixel. The perturbation $\delta$ is updated in each iteration as follows,

$$\delta_{t+1} = \delta_t + \varepsilon.sign(\nabla_x J(\theta, x+\delta, y)) \tag{5}$$

where, $J$ denotes the loss function, $y$ denotes the output class, $\nabla_x$ is a differential operator with respect to $x$, and $\varepsilon$ denotes the magnitude of perturbation.

## 4 AR-GAN FOR ADVERSARIAL ATTACK RESILIENT TRAFFIC SIGN CLASSIFICATION

In this section, the authors formally introduce the AR-GAN adversarial defense method. The traffic sign classification system of the AR-GAN includes a generator model and a classifier model obtained from the AR-GAN training framework. The generator is used to reconstruct any input images of traffic signs. This reconstruction process helps denoise the traffic sign images from adversarial noise. Subsequently, the classifier, trained on these reconstructed traffic sign images by the generator, helps classify the denoised traffic sign images.

### 4.1 AR-GAN Training Framework

The training framework to obtain the models in the traffic sign classification system of AR-GAN is depicted in Figure 1. First, the authors train a classifier model to classify unperturbed (i.e., legitimate or without attack) images in a traffic sign image dataset. In this study, the authors used a 9-layer deep residual learning architecture known as ResNet9 [10]. Once the classifier is trained to classify unperturbed traffic sign images with acceptable accuracy, the authors call it Classifier #1, which is used later to select the best generator model. Next, the authors train a set of GANs based on the WGAN architecture with a gradient penalty using the same unperturbed traffic sign images from the dataset. Once the GANs are trained, the authors utilize Classifier #1 to select the best generator model from the set of trained GANs that would yield the highest classification accuracy on the reconstructed traffic sign images of the test dataset. Then, the authors use the selected generator model to reconstruct all the unperturbed traffic sign images in the dataset. In an ideal case, if the GANs are trained to a point when the reconstructed traffic sign images look identical to the unperturbed traffic sign images, Classifier # 1 should be able to classify the reconstructed traffic sign images with similar accuracy to that of the unperturbed traffic sign images. However, if this level of accuracy is not achieved, the authors retrain Classifier #1 on the reconstructed traffic sign images to achieve better accuracy. The authors call this retrained classifier model Classifier #2. Finally, the AR-GAN traffic sign classification system is built with the best generator model, which reconstructs and denoises any input traffic sign images, and Classifier #2, which classifies these reconstructed images.

### 4.2 AR-GAN Classifier Model

The classifiers in the AR-GAN method are based on the ResNet9 architecture. ResNet9 is a 9-layer deep NN, including eight convolutional layers and one linear layer. The authors chose the ResNet9 architecture for developing the classifier models in this study because of the advantages of deep residual NNs over other CNNs and ResNet9's lightweight nature. Figure 2 presents the model architecture of the classifier used in the AR-GAN method, assuming the input traffic sign images have three color channels, i.e., red, green, and blue, each with $32 \times 32$ pixels. In Figure 2, the output dimension of each layer is presented as $C \times H \times W$, where C denotes the number of feature maps (for the input layer, it represents the number of color channels), and H and W denote the height and width of an output feature map, respectively.

### 4.3 AR-GAN Generator and Discriminator Models

GANs, first introduced by Goodfellow et al. [6], consist of two NNs, known as the generator ($G$) and the discriminator ($D$). $G: \mathbb{R}^k \rightarrow \mathbb{R}^n$ takes a low-dimensional input latent vector $z \in \mathbb{R}^k$ and maps it to a higher dimensional sample space of $x \in \mathbb{R}^n$. The discriminator, $D$ is a binary classifier that distinguishes real samples from fake (i.e., generated) samples. $G$ and $D$ are trained in tandem to optimize the following min-max loss function defined in [6],

$$\min_{G} \max_{D} V(D, G) = \mathrm{E}_{x \sim P_r(x)}[\log D(x)] + \mathrm{E}_{z \sim P_g(z)}\left[\log\left(1 - D(G(z))\right)\right] \quad (6)$$

where, $P_r(x)$ and $P_g(z)$ denote the real sample distribution and the generated sample distribution, respectively, and E denotes the expected value. The optimal GAN is obtained when these two distributions become the same. However, training GANs to optimality is difficult due to issues such as mode collapses and vanishing gradients. To resolve this, Arjovsky et al. [1] proposed a variant of the GAN, known as the Wasserstein GAN (WGAN), that utilizes the concepts of Wasserstein distance and Kantorovich-Rubinstein duality [30], with an alternative loss function given by,

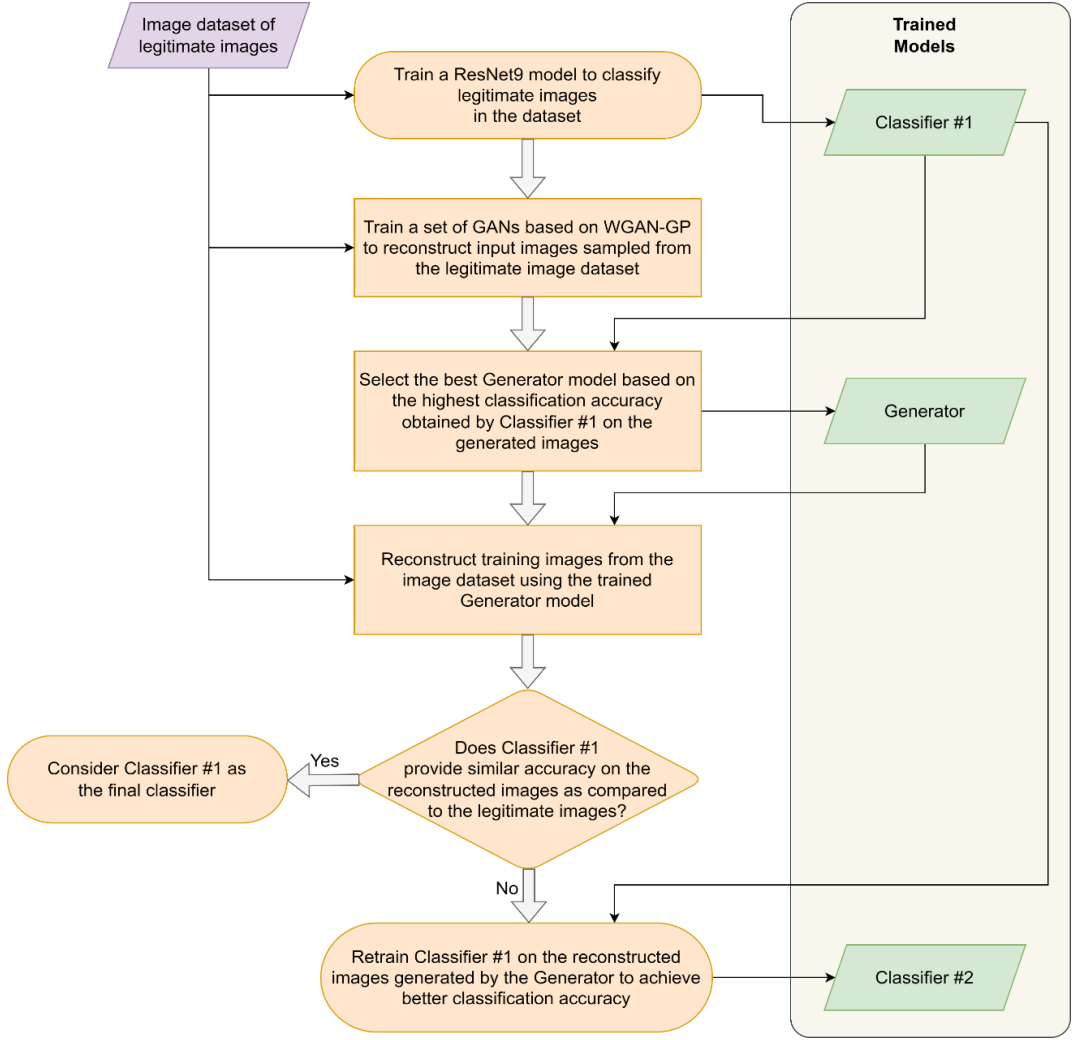

Figure 1: AR-GAN training framework.

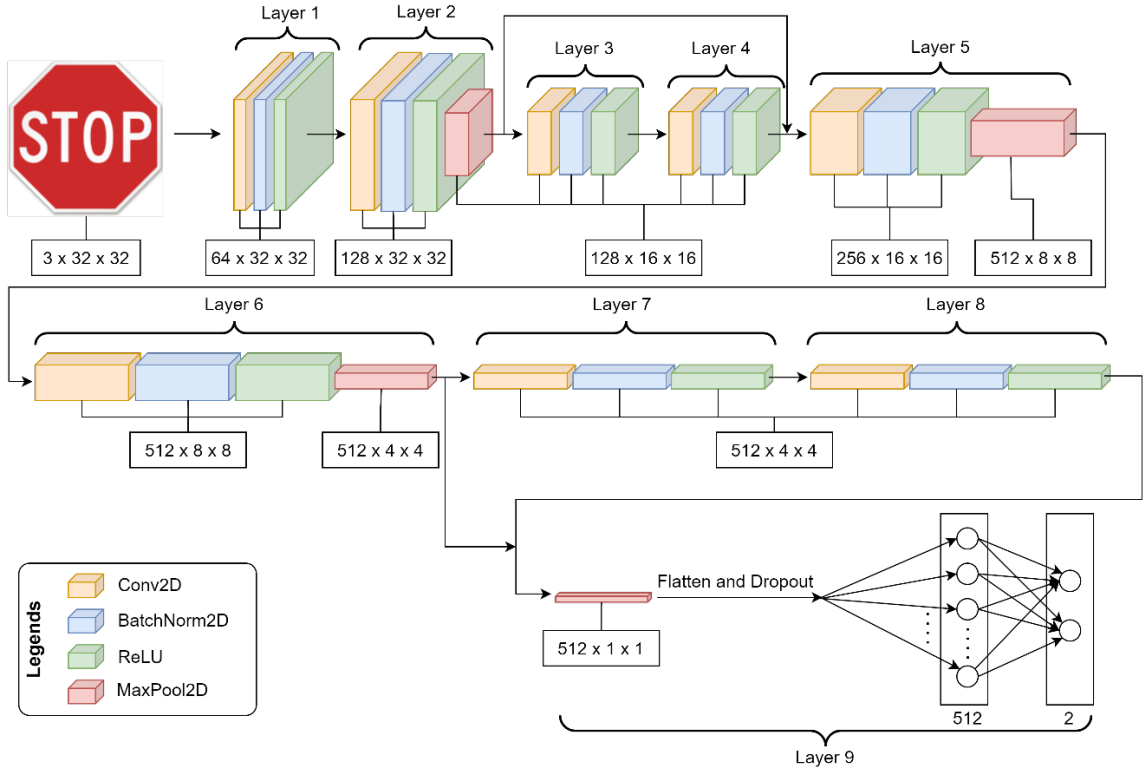

Figure 2: AR-GAN classifier based on the ResNet9 architecture.

$$\min_{G} \max_{D} V_W(D, G) = \mathrm{E}_{x \sim P_r(x)}[\log D(x)] - \mathrm{E}_{z \sim P_g(z)}\left[\log\left(D(G(z))\right)\right] \tag{7}$$

WGAN also removed the sigmoid function from the discriminator of the original GAN proposed in [6] to interpret the output of $D$ in terms of probability to indicate how "real" the generated images are. However, WGAN still suffered from some convergence issues due to the hard constraints set by the weight clipping method to enforce the Lipschitz condition. To address these issues, Gulrajani et al. [8] proposed an improved version known as WGAN with gradient penalty (WGAN-GP). WGAN-GP utilizes a soft version of the constraints by penalizing the model if the norm of the gradient deviates from its target norm value of 1 to meet the Lipschitz condition as follows [8],

$$\min_{G} \max_{D} V_W(D, G) = \mathrm{E}_{x \sim P_r(x)}[\log D(x)] - \mathrm{E}_{z \sim P_g(z)}\left[\log\left(D(G(z))\right)\right] + \lambda \mathrm{E}_{\hat{x} \sim P_{\hat{x}}}[\|\nabla_{\hat{x}} D(\hat{x})\|_2 - 1]^2 \tag{8}$$

Here, $\lambda$ is set to 10, $\nabla_{\hat{x}}$ is a differential operator with respect to $\hat{x}$, and $\hat{x}$ is sampled from $x$ and $G(x)$ using the following linear equation,

$$\hat{x} = tG(x) + (1-t)x \tag{9}$$

where, $t$ is uniformly sampled between 0 and 1, i.e., $0 \leq t \leq 1$. The WGAN-GP, proposed in [8], removed the batch normalization steps from the discriminator as it affected the effectiveness of the gradient penalty.

However, to be able to reconstruct an input image with minimum reconstruction error, further extension is needed. The authors in [36] proposed a simple extension to achieve this by optimizing the input latent vector $z$ that will be fed to the generator to reconstruct an input image $x$, which the authors adopt in the AR-GAN method. The optimal latent vector $z^*$ for reconstructing an input image $x$ is obtained by solving the optimization problem given by,

$$z^* = \arg\min_z \|G(z) - x\|_2^2 \tag{10}$$

Equation (10) is solved in a gradient descent-based iterative approach. Because of its non-convex nature, the authors in [36] utilized a fixed number of gradient descent steps along with a given number of random initializations of the latent vector $z$.

Figures 3 and 4 present the architectures of the generator and the discriminator used in the AR-GAN method. The generator architecture is based on the DCGAN architecture [31], whereas the discriminator architecture is kept the same as the WGAN architecture. In Figures 3 and 4, the output dimension of each layer is presented as C × H × W, where C denotes the number of feature maps, and H and W denote the height and width of each feature map, respectively.

Finally, the AR-GAN traffic sign classification system consists of the trained generator, the classifier, and the optimizer that optimizes the input latent vector applied to the generator to reconstruct an input traffic sign image. Figure 5 presents the AR-GAN traffic sign classification system with a flow diagram.

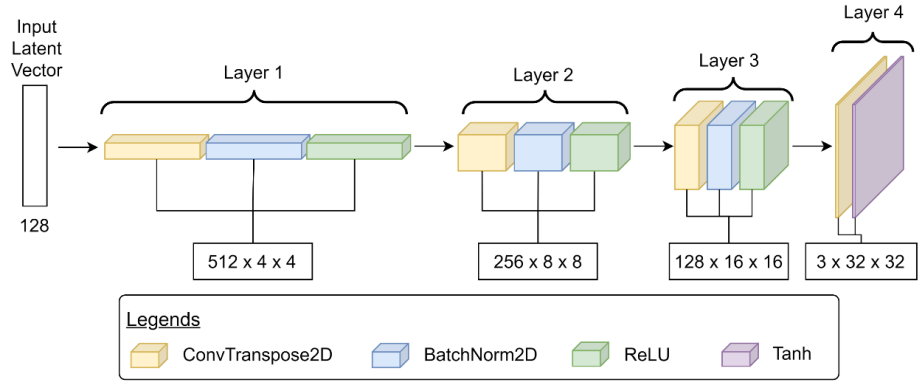

Figure 3: AR-GAN generator based on the DCGAN architecture.

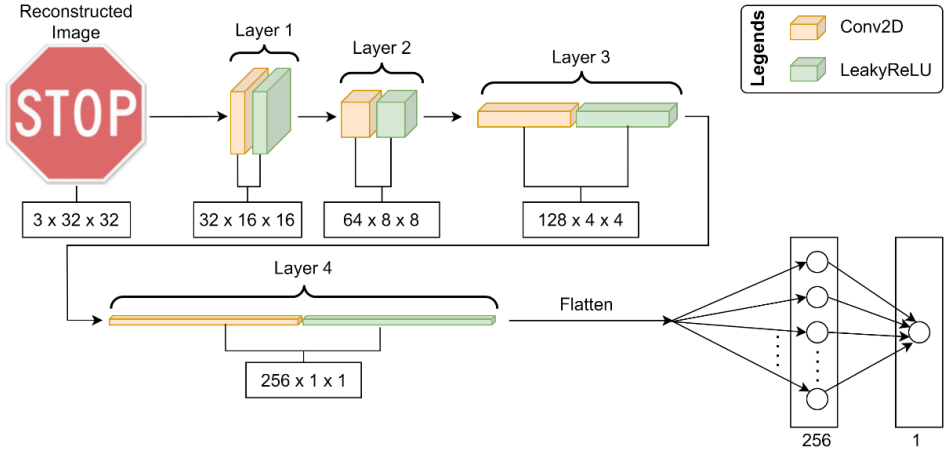

Figure 4: AR-GAN discriminator based on the WGAN architecture.

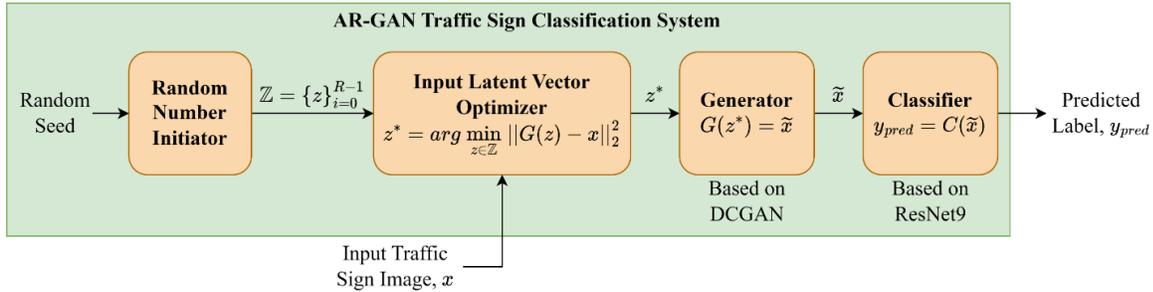

Figure 5: AR-GAN traffic sign classification system.

## 5 EVALUATION METHOD

This section discusses the evaluation approach utilized in this study, outlining the traffic sign dataset and the traditional preprocessing-based defense methods the authors employed for comparison.

### 5.1 Real-World Traffic Sign Dataset

The authors reviewed the existing literature to select a comprehensive US traffic sign dataset and found the LISA traffic sign dataset [25] the most appropriate since it contains data collected from the real world. The LISA dataset covers 49 types of US traffic signs with 7,855 annotations on 6,610 frames. The traffic sign images in the LISA dataset were extracted from video frames captured by multiple vehicles' dashboard cameras while the vehicles roamed around San Diego, California. The LISA video frames exhibit varying resolutions, ranging from 640 × 480 to 1024 × 522 pixels. The traffic sign annotations have dimensions spanning from 6 × 6 to 167 × 168 pixels and include both color and grayscale images.

However, the LISA dataset does not contain enough images for each type of traffic sign to train GAN models. Therefore, the authors created a subset of the LISA dataset containing two traffic sign classes with the highest number of images, i.e., STOP signs and SPEED LIMIT signs. The original LISA dataset contains different types of SPEED LIMIT signs, such as 15, 25, 30, 35, 40, 45, 50, and 65 miles per hour (mph) signs. The authors combined all these SPEED LIMIT signs into one class to create a balanced dataset. The subset of the LISA dataset used in this study for evaluation of the AR-GAN method contains a total of 1,562 traffic sign images, including 805 images under the STOP sign class and 757 images under the SPEED LIMIT sign class. The authors applied cropping and resizing to ensure that all the images in the dataset have the same dimension, i.e., each image has three channels for red, green, and blue colors, and each channel contains 32 × 32 pixels. Figure 6 presents some sample images from the dataset used in this study. As observed from the figure, the images are not very clean and contain some noise, which makes them even harder to classify under adversarial perturbations.

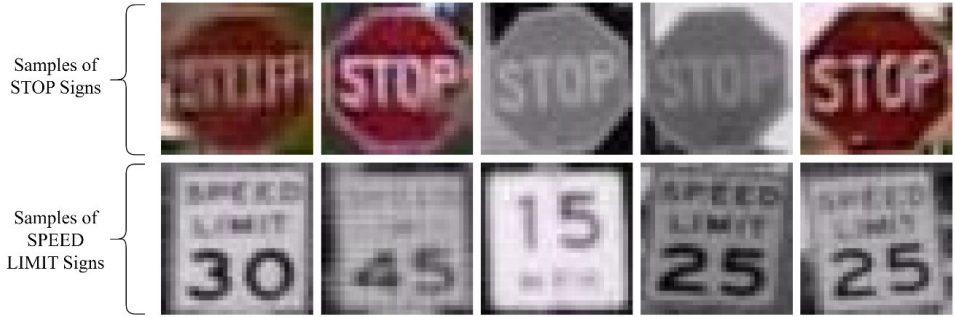

Figure 6: Image samples from the LISA traffic sign dataset.

## 5.2 Traditional Preprocessing-Based Defense Methods

The authors selected several traditional preprocessing-based defense methods that can be used as benchmarks to compare with the classification performance of the AR-GAN traffic sign classification system. Table 1 highlights the traditional preprocessing-based adversarial attack defense methods considered in this study. Figure 7 shows the effect of these defense methods, including the AR-GAN method, on a set of sample traffic sign images used in this study.

Table 1: Traditional Preprocessing-based Defense Methods Used in This Study

| Defense Method | Method Highlights | Method Settings Used in This Study |
|---|---|---|
| Gaussian Augmentation [7] | • Applies random noise to every pixel of an input image to grow robustness against adversarial attacks<br>• Applies independent and identically distributed noise sampled from a zero-mean Gaussian distribution, $\mathcal{N}(0, \sigma^2)$, to each pixel of an input image | Standard deviation $\sigma$ is set to 1, as recommended in [7] |
| JPEG Compression [21] | Eliminates high-frequency components from an input image that may cause the DNN-based classifiers to misclassify the image under an adversarial attack | Compressed image quality is set to 50%, as recommended in [21] |
| Feature Squeezing [38] | • Reduces the depth of each pixel's color bits of an input image that is necessary to represent the color value of that pixel<br>• Transforms diverse feature vectors from the original space into more similar samples | Bit depth value is set to 4, as recommended in [38] |

| Defense Method | Method Highlights | Method Settings Used in This Study |
|---|---|---|
| Median Smoothing [38] | Reduces differences among the pixel values of an input image by moving a sliding window across the image while the window's center pixel value is replaced by its neighboring pixels' median value | Size of the sliding window is set to 3 × 3 |

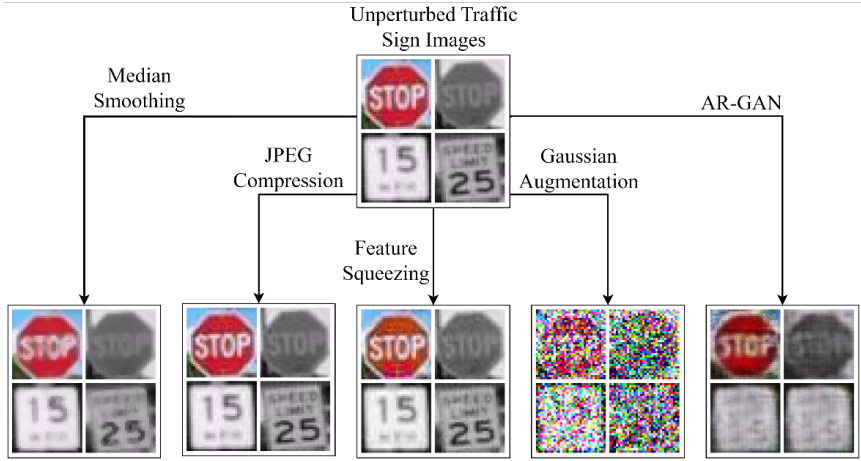

Figure 7: Examples of preprocessed images using different defense methods.

## 6 ANALYSIS AND RESULTS

This section presents the analysis and results based on the AR-GAN method and compares them with the traditional preprocessing-based defense methods. The authors divide the evaluation scenarios into two categories, i.e., (i) evaluation on unperturbed traffic sign images and adversarial images, and (ii) evaluation under different perturbation magnitudes. The dataset was split into three subsets, i.e., train set (containing 60% of the images), validation set (containing 20% of the images), and test set (containing the remaining 20% of the images), after applying random shuffling on all the images on the dataset. The same train, validation, and test sets were used for all the evaluation scenarios to present a fair comparison among the different defense methods used in this study.

The AR-GAN traffic sign classification system utilizes a gradient descent-based optimization to determine the input latent vector that would minimize the reconstruction error for an input image. The authors conducted a sensitivity analysis of the AR-GAN traffic sign classification system's end-to-end delay (i.e., the time required to perform all the steps shown in Figure 5) and classification accuracy with respect to the number of gradient descent steps and the number of random initializations (see Figure 8). For this analysis, the authors used all the images in the test set and considered the average values of delay and accuracy. As observed from Figure 8(a), the end-to-end delay per image increases with the increase in the number of gradient descent steps as well as the number of random initializations. Liu and Deng [20] reported the average delay for human drivers in recognizing traffic signs to be ranging from 0.5 to 2.0 seconds, which sets a limit to the number of gradient descent steps as well as the number of random initializations we can allow. In addition, the lower the end-to-end delay is, the better it is for real-world implementation in AVs. On the other hand, as observed from Figure 8(b), the classification accuracy improves initially while increasing the number of random initializations, and then it starts to drop. The highest traffic sign classification accuracy was achieved for 2,250 gradient descent steps and 20 random initializations, which resulted in an end-to-end delay of 0.6 seconds. The authors considered this delay feasible for real-world implementation as it is much lower than the human drivers' average traffic sign recognition delay reported in [20].

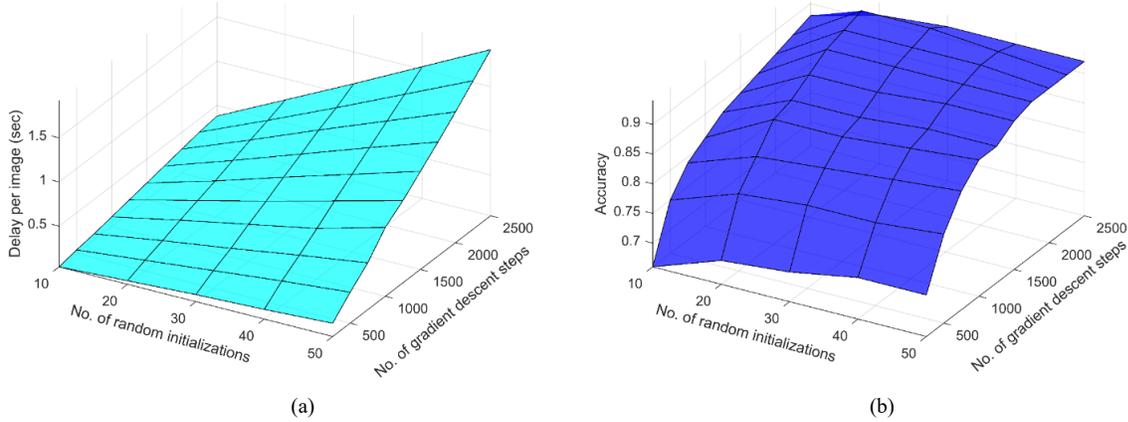

Figure 8: Sensitivity analysis results of average (a) delay per image and (b) classification accuracy.

The authors used Pytorch packages [42] to implement the GAN and the classifier models, and the Adversarial Robustness Toolbox [43] to implement the attack models and traditional preprocessing-based defense methods. The source codes are provided in GitHub [35]. All the NN models were trained using Nvidia Tesla A100 GPUs available in the Palmetto Cluster nodes at Clemson University [41]. These GPUs have a capacity of 312 trillion floating point operations per second (TFLOPS) [44]. One of the recent in-vehicle computational units developed by NVIDIA is the NVIDIA Drive Thor, offering a GPU-based computational capacity of 2,000 TFLOPS [45], which is well above the capacity of the A100 GPUs utilized in this study for training the NN models. Besides, the in-vehicle computational units should only be responsible for running pretrained models for traffic sign image classification, whereas the training task can take place separately beforehand. Thus, the models developed under the AR-GAN method are considered feasible to be implemented in real-world AVs in terms of in-vehicle computational capacity.

### 6.1 Evaluation on Unperturbed and Adversarial Traffic Sign Images

Table 2 presents the evaluation results obtained using unperturbed traffic sign images. Table 3 and 4 present the evaluation results obtained using black-box and white-box adversarial images, respectively. The adversarial images were generated under FGSM, DeepFool, C&W, and PGD attacks. The authors used precision, recall, F1-score, and accuracy for traffic sign classification performance comparison among the different defense methods. Among the performance metrics, accuracy was calculated globally for all the images in the test set, whereas the other metrics were calculated for each class, and then a weighted average was taken.

As shown in Table 2, all the defense methods considered in this study achieved high precision, recall, F1-score, and accuracy on the unperturbed images. This proves that the classifier models used in all the defense methods in this study are well-trained to accurately classify the traffic sign images of the dataset. Although the AR-GAN method achieved about 95% classification accuracy on the unperturbed images, it was lower than the most other methods. This difference in performance is because, unlike the other preprocessing-based defense methods that transform or modify an input image, the AR-GAN completely reconstructs any input images. Gaussian augmentation achieved the second-lowest classification performance compared to the other defense methods, which is also expected because this defense method itself adds some Gaussian noise to the images as part of its adversarial defense strategy.

Table 2: Comparison of Defense Methods on Unperturbed Images

| Defense Method | Precision | Recall | F1-score | Accuracy |
| --- | --- | --- | --- | --- |
| Gaussian Augmentation | 95.1% | 94.9% | 94.9% | 94.9% |
| JPEG Compression | 99.7% | 99.7% | 99.7% | 99.7% |
| Feature Squeezing | 99.7% | 99.7% | 99.7% | 99.7% |
| Median smoothing | 98.8% | 98.7% | 98.7% | 98.7% |
| **AR-GAN** | **95.0%** | **94.9%** | **94.9%** | **94.9%** |

Under black-box conditions, the inherent limitations faced by attackers due to the absence of direct access to the architecture and weights of a target classifier likely lead to less effective adversarial attacks [32]. As a result, it is observed from Table 3 that all the defense methods, including the AR-GAN, yield classification performance close to those observed for unperturbed images presented in Table 2.

Table 3: Comparison of Defense Methods on Black-box Adversarial Images

| Attack Type | Defense Method | Precision | Recall | F1-score | Accuracy |
| --- | --- | --- | --- | --- | --- |
| FGSM | Gaussian Augmentation | 94.3% | 94.3% | 94.3% | 94.3% |
| | JPEG Compression | 99.7% | 99.7% | 99.7% | 99.7% |
| | Feature Squeezing | 99.4% | 99.4% | 99.4% | 99.4% |
| | Median smoothing | 97.6% | 97.4% | 97.4% | 97.4% |
| | **AR-GAN** | **94.6%** | **94.6%** | **94.6%** | **94.6%** |
| DeepFool | Gaussian Augmentation | 94.3% | 94.3% | 94.3% | 94.3% |
| | JPEG Compression | 99.7% | 99.7% | 99.7% | 99.7% |
| | Feature Squeezing | 99.7% | 99.7% | 99.7% | 99.7% |
| | Median smoothing | 98.8% | 98.7% | 98.7% | 98.7% |
| | **AR-GAN** | **94.8%** | **94.6%** | **94.6%** | **94.6%** |
| C&W | Gaussian Augmentation | 95.2% | 94.9% | 94.9% | 94.9% |
| | JPEG Compression | 99.7% | 99.7% | 99.7% | 99.7% |
| | Feature Squeezing | 99.7% | 99.7% | 99.7% | 99.7% |
| | Median smoothing | 98.8% | 98.7% | 98.7% | 98.7% |
| | **AR-GAN** | **95.6%** | **95.5%** | **95.5%** | **95.5%** |
| PGD | Gaussian Augmentation | 94.4% | 94.3% | 94.2% | 94.3% |
| | JPEG Compression | 99.7% | 99.7% | 99.7% | 99.7% |
| | Feature Squeezing | 99.7% | 99.7% | 99.7% | 99.7% |
| | Median smoothing | 98.8% | 98.7% | 98.7% | 98.7% |
| | **AR-GAN** | **93.4%** | **93.3%** | **93.3%** | **93.3%** |

On the contrary, attackers can fine-tune their attack models using knowledge of the classifier's architecture and weights in a white-box attack. The novelty of the AR-GAN method lies in its consistent resilience to both black-box and white-box attacks. However, the performance gap between the traditional defense methods and the AR-GAN method becomes apparent under white-box attacks, as depicted in Table 4.

Table 4: Comparison of Defense Methods on White-box Adversarial Images

| Attack Type | Defense Method | Precision | Recall | F1-score | Accuracy |
|---|---|---|---|---|---|
| FGSM | Gaussian Augmentation | 81.1% | 80.5% | 80.4% | 80.5% |
| | JPEG Compression | 75.4% | 75.4% | 75.4% | 75.4% |
| | Feature Squeezing | 68.7% | 68.7% | 68.7% | 68.7% |
| | Median smoothing | 68.5% | 68.4% | 68.3% | 68.4% |
| | **AR-GAN** | **92.8%** | **92.7%** | **92.6%** | **92.7%** |
| DeepFool | Gaussian Augmentation | 74.0% | 73.8% | 73.8% | 73.8% |
| | JPEG Compression | 62.0% | 61.3% | 60.8% | 61.3% |
| | Feature Squeezing | 32.3% | 32.3% | 32.3% | 32.3% |
| | Median smoothing | 42.1% | 42.8% | 41.6% | 42.8% |
| | **AR-GAN** | **92.7%** | **92.7%** | **92.7%** | **92.7%** |
| C&W | Gaussian Augmentation | 79.0% | 78.6% | 78.5% | 78.6% |
| | JPEG Compression | 62.9% | 61.7% | 60.8% | 61.7% |
| | Feature Squeezing | 26.2% | 26.2% | 26.2% | 26.2% |
| | Median smoothing | 38.1% | 38.7% | 37.9% | 38.7% |
| | **AR-GAN** | **93.4%** | **93.3%** | **93.3%** | **93.3%** |
| PGD | Gaussian Augmentation | 75.6% | 75.4% | 75.4% | 75.4% |
| | JPEG Compression | 58.8% | 58.5% | 58.1% | 58.5% |
| | Feature Squeezing | 44.1% | 44.1% | 44.0% | 44.1% |
| | Median smoothing | 48.9% | 48.9% | 48.9% | 48.9% |
| | **AR-GAN** | **91.7%** | **91.7%** | **91.7%** | **91.7%** |

The FGSM attack was implemented with an $\varepsilon = 0.1$ perturbation magnitude, as recommended by Ye and Zhu [39]. The FGSM attack is not as powerful as the other attacks used in this study. As observed from Table 4, the performance metrics of the traditional preprocessing-based defense methods ranged from approximately 68% to 81%. However, the AR-GAN method improved the overall performance by about 10-12% compared to the second-best defense method, i.e., the Gaussian augmentation.

As observed from Table 4, the $\ell_2$-norm DeepFool attack, with a perturbation magnitude of $\varepsilon = 0.1$, was more effective than the FGSM attack in degrading the classification performance. Feature squeezing and median smoothing performed the worst among all the defense methods. Gaussian augmentation was able to achieve about 74% classification accuracy, which was outperformed by the AR-GAN method with a classification accuracy of about 93%.

The $\ell_2$-norm C&W attack was performed using a learning rate of 0.01 with a maximum of 10 iterations. As seen from Table 4, feature squeezing and median smoothing provided the worst traffic sign classification accuracies among all the defense methods. Gaussian augmentation achieved about 79% classification accuracy. However, the AR-GAN method outperformed all the other defense methods with a 93% classification accuracy under the C&W attack.

As observed from Table 4, the $\ell_2$-norm PGD attack, with a maximum iteration number of 100 and a perturbation magnitude of $\varepsilon = 0.1$, caused the traffic sign classification accuracies to drop under 60% for all the traditional defense methods, except for the Gaussian augmentation preprocessing, which achieved about 75% accuracy. Again, the AR-GAN method outperformed all the traditional defense methods with a classification accuracy of approximately 92%.

## 6.2 Evaluation under Different Perturbation Magnitudes

To evaluate how well the AR-GAN method performs under different perturbation magnitudes, the authors varied $\varepsilon$ from 0.05 to 0.2 with a 0.05 step size for the FGSM, DeepFool, and PGD attacks following the previous studies [4,14,27].

Figure 9 presents the results of this evaluation under different perturbation magnitudes. From Figures 9(a) to 9(c), it is observed that all the defense methods, including the AR-GAN, were able achieve above 93% classification accuracy under the black-box FGSM, DeepFool, and PGD attacks with varied $\varepsilon$. However, the accuracies of the traditional preprocessing-based defense methods dropped abruptly as the authors increased $\varepsilon$ of the white-box FGSM and PGD attacks, as observed from Figures 9(d) and 9(f), respectively. In Figure 9(e), it is observed that these drops in traffic sign classification performance happen gradually for the traditional preprocessing-based defense methods under the white-box DeepFool attack. However, the AR-GAN method achieved above 90% classification accuracy in all these cases, except under the PGD attacks with $\varepsilon = 0.15$ and 0.2, where its accuracy dropped to about 87%. This consistency in traffic sign classification performance is achievable with the AR-GAN method because the generator in the AR-GAN method was trained to generate samples close to the unperturbed traffic sign images' distribution. Thus, the AR-GAN method developed in this study can effectively denoise the traffic sign images by reconstructing them with a generator trained on the unperturbed traffic sign images without any prior knowledge of adversarial attack types and samples.

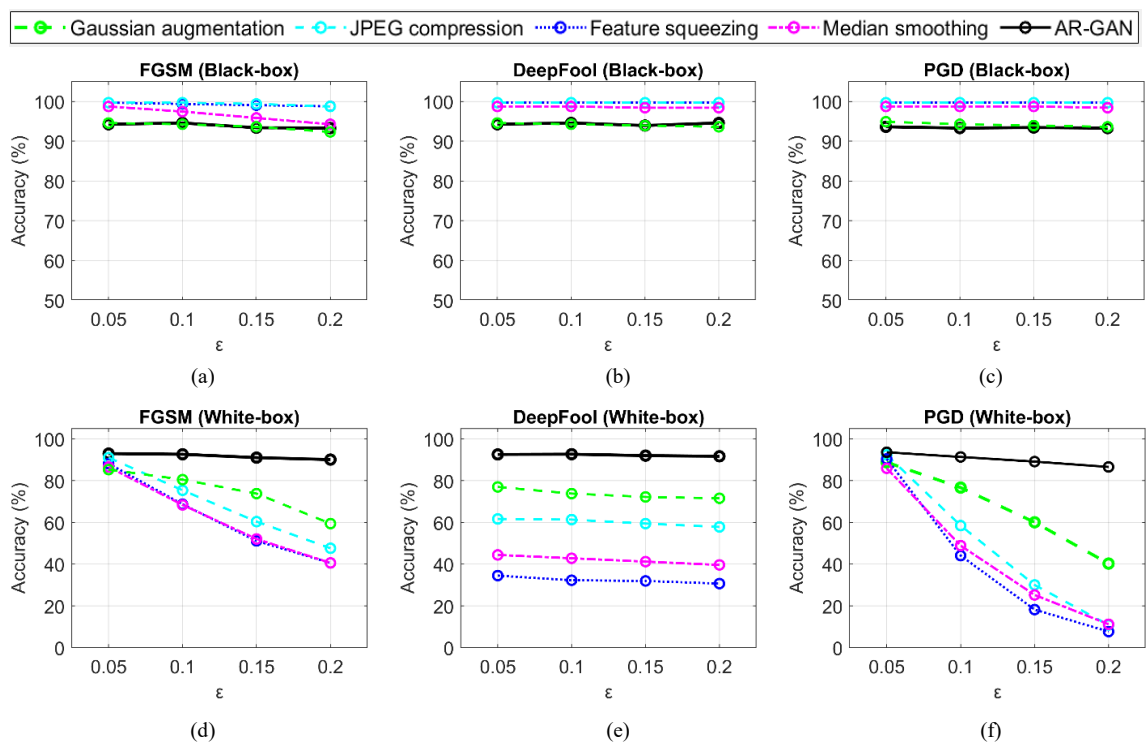

Figure 9: Comparison of defense methods across varying perturbation magnitudes, $\varepsilon$, for: (a) FGSM (black-box), (b) DeepFool (black-box), (c) PGD (black-box), (d) FGSM (white-box), (e) DeepFool (white-box), and (f) PGD (white-box) attacks.

## 7 CONCLUSIONS

In this study, the authors developed a GAN-based adversarial defense method for an AV traffic sign classification system. The authors provided a framework for training a generator model and a classifier model, which comprise the attack-resilient traffic sign classification system of the AR-GAN method. The generator model was trained using a WGAN-GP-based loss function with a DCGAN architecture. The discriminator, used to support the training of the generator, was based on the WGAN architecture. The classifier was trained based on the ResNet9 architecture. All these models were trained only with unperturbed (i.e., legitimate) traffic sign images to ensure that any adversarial attacks are unknown to the trained models.

The traffic sign classification system of the AR-GAN defense method utilizes a generator-based image reconstruction to eliminate adversarial perturbations from an input traffic sign image. Once reconstructed, the image is fed to the classifier to identify the type of traffic sign. The authors evaluated the AR-GAN traffic sign classification system against widely used black-box and white-box adversarial attacks, such as FGSM, DeepFool, C&W, and PGD attacks, and compared its performance with benchmark traditional adversarial defense methods, such as Gaussian augmentation, JPEG compression, feature squeezing, and median smoothing.

Under the black-box attacks considered in this study, the AR-GAN and the other traditional defense methods exhibited classification performance comparable to that on the unperturbed images. However, when encountered with white-box attacks, which assume that the attackers possess complete knowledge of the classification models, the AR-GAN method demonstrated superior resilience, outperforming all the traditional defense methods evaluated in this study. In addition, for the white-box adversarial images, the AR-GAN was able to consistently achieve high traffic sign classification performance under various adversarial perturbation magnitudes, whereas the performance for the other traditional defense methods dropped abruptly at increased perturbation levels. This shows the potential of the AR-GAN method to be deployed as a robust AV traffic sign classification system to achieve resiliency against various types of adversarial attacks.

## ACKNOWLEDGMENTS

This work is based upon the work partially supported by the National Center for Transportation Cybersecurity and Resiliency (TraCR) (a U.S. Department of Transportation National University Transportation Center) headquartered at Clemson University, Clemson, South Carolina, USA. Any opinions, findings, conclusions, and recommendations expressed in this material are those of the author(s) and do not necessarily reflect the views of TraCR, and the U.S. Government assumes no liability for the contents or use thereof.

[10] Kaiming He, Xiangyu Zhang, Shaoqing Ren, and Jian Sun. 2016. Deep Residual Learning for Image Recognition. 2016. 770–778. . Retrieved July 7, 2023 from https://openaccess.thecvf.com/content_cvpr_2016/html/He_Deep_Residual_Learning_CVPR_2016_paper.html

[11] Guoqing Jin, Shiwei Shen, Dongming Zhang, Feng Dai, and Yongdong Zhang. 2019. APE-GAN: Adversarial Perturbation Elimination with GAN. In *ICASSP 2019 - 2019 IEEE International Conference on Acoustics, Speech and Signal Processing (ICASSP)*, May 2019. 3842–3846. . https://doi.org/10.1109/ICASSP.2019.8683044

[12] Abdulrahman Kerim and Mehmet Önder Efe. 2021. Recognition of Traffic Signs with Artificial Neural Networks: A Novel Dataset and Algorithm. In *2021 International Conference on Artificial Intelligence in Information and Communication (ICAIIC)*, April 2021. 171–176. . https://doi.org/10.1109/ICAIIC51459.2021.9415238

[13] Samer Y. Khamaiseh, Derek Bagagem, Abdullah Al-Alaj, Mathew Mancino, and Hakam W. Alomari. 2022. Adversarial Deep Learning: A Survey on Adversarial Attacks and Defense Mechanisms on Image Classification. *IEEE Access* 10, (2022), 102266–102291. https://doi.org/10.1109/ACCESS.2022.3208131

[14] Zadid Khan, Mashrur Chowdhury, and Sakib Mahmud Khan. 2022. A hybrid defense method against adversarial attacks on traffic sign classifiers in autonomous vehicles. *arXiv preprint arXiv:2205.01225* (2022).

[15] Mohammed Qader Kheder and Aree Ali Mohammed. 2023. Improved traffic sign recognition system (itsrs) for autonomous vehicle based on deep convolutional neural network. *Multimed Tools Appl* (May 2023). https://doi.org/10.1007/s11042-023-15898-6

[16] Taeksoo Kim, Moonsu Cha, Hyunsoo Kim, Jung Kwon Lee, and Jiwon Kim. 2017. Learning to Discover Cross-Domain Relations with Generative Adversarial Networks. In *Proceedings of the 34th International Conference on Machine Learning*, July 17, 2017. PMLR, 1857–1865. . Retrieved July 10, 2023 from https://proceedings.mlr.press/v70/kim17a.html

[17] Pranpaveen Laykaviriyakul and Ekachai Phaisangittisagul. 2023. Collaborative Defense-GAN for protecting adversarial attacks on classification system. *Expert Systems with Applications* 214, (March 2023), 118957. https://doi.org/10.1016/j.eswa.2022.118957

[18] Hailiang Li, Bin Zhang, Yu Zhang, Xilin Dang, Yuwei Han, Linfeng Wei, Yijun Mao, and Jian Weng. 2021. A defense method based on attention mechanism against traffic sign adversarial samples. *Information Fusion* 76, (December 2021), 55–65. https://doi.org/10.1016/j.inffus.2021.05.005

[19] Xin Roy Lim, Chin Poo Lee, Kian Ming Lim, Thian Song Ong, Ali Alqahtani, and Mohammed Ali. 2023. Recent Advances in Traffic Sign Recognition: Approaches and Datasets. *Sensors* 23, 10 (January 2023), 4674. https://doi.org/10.3390/s23104674

[20] Kun Liu and Hongxing Deng. 2021. The Analysis of Driver's Recognition Time of Different Traffic Sign Combinations on Urban Roads via Driving Simulation. *Journal of Advanced Transportation* 2021, (August 2021), e8157293. https://doi.org/10.1155/2021/8157293

[21] Zihao Liu, Qi Liu, Tao Liu, Nuo Xu, Xue Lin, Yanzhi Wang, and Wujie Wen. 2019. Feature Distillation: DNN-Oriented JPEG Compression Against Adversarial Examples. In *2019 IEEE/CVF Conference on Computer Vision and Pattern Recognition (CVPR)*, June 2019. 860–868. . https://doi.org/10.1109/CVPR.2019.00095

[22] Aleksander Madry, Aleksandar Makelov, Ludwig Schmidt, Dimitris Tsipras, and Adrian Vladu. 2017. Towards Deep Learning Models Resistant to Adversarial Attacks. *arXiv.org*. Retrieved July 6, 2023 from https://arxiv.org/abs/1706.06083v4

[23] Reek Majumder, Sakib Mahmud Khan, Fahim Ahmed, Zadid Khan, Frank Ngeni, Gurcan Comert, Judith Mwakalonge, Dimitra Michalaka, and Mashrur Chowdhury. 2021. Hybrid Classical-Quantum Deep Learning Models for Autonomous Vehicle Traffic Image Classification Under Adversarial Attack. https://doi.org/10.48550/arXiv.2108.01125

[24] Enrique Marti, Miguel Angel de Miguel, Fernando Garcia, and Joshue Perez. 2019. A Review of Sensor Technologies for Perception in Automated Driving. *IEEE Intelligent Transportation Systems Magazine* 11, 4 (2019), 94–108. https://doi.org/10.1109/MITS.2019.2907630

[25] Andreas Mogelmose, Mohan Manubhai Trivedi, and Thomas B Moeslund. 2012. Vision-based traffic sign detection and analysis for intelligent driver assistance systems: Perspectives and survey. *IEEE transactions on intelligent transportation systems* 13, 4 (2012), 1484–1497.

[26] Seyed-Mohsen Moosavi-Dezfooli, Alhussein Fawzi, and Pascal Frossard. 2016. Deepfool: a simple and accurate method to fool deep neural networks. 2016. 2574–2582. .

[27] Rangeet Pan, Md Johirul Islam, Shibbir Ahmed, and Hridesh Rajan. 2019. Identifying Classes Susceptible to Adversarial Attacks. https://doi.org/10.48550/arXiv.1905.13284

[28] Raji Pandurangan, Samuel Manoharan Jayaseelan, Suresh Rajalingam, and Kandavalli Michael Angelo. 2023. A novel hybrid machine learning approach for traffic sign detection using CNN-GRNN. *Journal of Intelligent & Fuzzy Systems* 44, 1 (January 2023), 1283–1303. https://doi.org/10.3233/JIFS-221720

[29] Nicolas Papernot, Patrick McDaniel, Xi Wu, Somesh Jha, and Ananthram Swami. 2016. Distillation as a Defense to Adversarial Perturbations Against Deep Neural Networks. In *2016 IEEE Symposium on Security and Privacy (SP)*, May 2016. 582–597. . https://doi.org/10.1109/SP.2016.41

[30] Svetlozar Todorov Rachev. 1990. Duality theorems for Kantorovich-Rubinstein and Wasserstein functionals. (1990).

[31] Alec Radford, Luke Metz, and Soumith Chintala. 2016. Unsupervised Representation Learning with Deep Convolutional Generative Adversarial Networks. https://doi.org/10.48550/arXiv.1511.06434

[32] Kui Ren, Tianhang Zheng, Zhan Qin, and Xue Liu. 2020. Adversarial Attacks and Defenses in Deep Learning. *Engineering* 6, 3 (March 2020), 346–360. https://doi.org/10.1016/j.eng.2019.12.012

[33] Andrew Ross and Finale Doshi-Velez. 2018. Improving the Adversarial Robustness and Interpretability of Deep Neural Networks by Regularizing Their Input Gradients. *Proceedings of the AAAI Conference on Artificial Intelligence* 32, 1 (April 2018). https://doi.org/10.1609/aaai.v32i1.11504

[34] Yassmina Saadna and Ali Behloul. 2017. An overview of traffic sign detection and classification methods. *Int J Multimed Info Retr* 6, 3 (September 2017), 193–210. https://doi.org/10.1007/s13735-017-0129-8

[35] M Sabbir Salek. 2023. msabbirsalek/AR-GAN. Retrieved July 12, 2023 from https://github.com/msabbirsalek/AR-GAN

[36] Pouya Samangouei, Maya Kabkab, and Rama Chellappa. 2018. Defense-GAN: Protecting Classifiers Against Adversarial Attacks Using Generative Models. https://doi.org/10.48550/arXiv.1805.06605

[37] Safat B. Wali, Majid A. Abdullah, Mahammad A. Hannan, Aini Hussain, Salina A. Samad, Pin J. Ker, and Muhamad Bin Mansor. 2019. Vision-Based Traffic Sign Detection and Recognition Systems: Current Trends and Challenges. *Sensors* 19, 9 (January 2019), 2093. https://doi.org/10.3390/s19092093

[38] Weilin Xu, David Evans, and Yanjun Qi. 2018. Feature Squeezing: Detecting Adversarial Examples in Deep Neural Networks. In *Proceedings 2018 Network and Distributed System Security Symposium*, 2018. . https://doi.org/10.14722/ndss.2018.23198

[39] Nanyang Ye and Zhanxing Zhu. 2018. Bayesian Adversarial Learning. In *Advances in Neural Information Processing Systems*, 2018. Curran Associates, Inc. . Retrieved July 8, 2023 from https://proceedings.neurips.cc/paper_files/paper/2018/hash/586f9b4035e5997f77635b13cc04984c-Abstract.html

[40] Huan Zhang, Hongge Chen, Zhao Song, Duane Boning, Inderjit S. Dhillon, and Cho-Jui Hsieh. 2019. The Limitations of Adversarial Training and the Blind-Spot Attack. https://doi.org/10.48550/arXiv.1901.04684
18

# 8  HISTORY DATES